\documentclass[final]{cvpr}

\usepackage{times}
\usepackage{epsfig}
\usepackage{graphicx}
\usepackage{amsmath}
\usepackage{amssymb}
\usepackage{caption}
\usepackage{subcaption}

\newcommand{\squeezeup}{\vspace{-2mm}}
\usepackage{adjustbox}
\usepackage{booktabs}       
\usepackage{nicefrac}       
\usepackage{microtype}      
\usepackage{multirow}   

\usepackage[pagebackref=true,breaklinks=true,colorlinks,bookmarks=false]{hyperref}

\def\cvprPaperID{6272} 
\def\confYear{CVPR 2021}

\begin{document}

\title{Rethinking Movie Genre Classification with Fine Grained Semantic Clustering}

\author{Edward Fish\\
University of Surrey\\

{\tt\small edward.fish@surrey.ac.uk}

\and
Jon Weinbren\\
University of Surrey \\
{\tt\small j.weinbren@surrey.ac.uk}

\and
Andrew Gilbert\\
University of Surrey \\

{\tt\small a.gilbert@surrey.ac.uk}

}
\maketitle

\begin{abstract}

Movie genre classification is an active research area in machine learning. However, due to the limited labels available, there can be large semantic variations between movies within a single genre definition. We expand these `coarse' genre labels by identifying `fine-grained' semantic information within the multi-modal content of movies. By leveraging pre-trained `expert' networks, we learn the influence of different combinations of modes for multi-label genre classification. Using a contrastive loss, we continue to fine-tune this `coarse' genre classification network to identify high-level intertextual similarities between the films across all genre labels. This leads to a more `fine-grained' and detailed clustering, based on semantic similarities while still retaining some genre information. Our approach is demonstrated on a newly introduced multi-modal 37,866,450 frame, 8,800 movie trailer dataset, MMX-Trailer-20, which includes pre-computed audio, location, motion, and image embeddings. 

\end{abstract}

\squeezeup
\squeezeup
\section{Introduction}


Genres are a useful classification device for condensing the content of a movie into an easy to understand contextual frame for the viewer. However, in the field of Film Theory, genre is not perceived as a reliable descriptive label for a number of reasons. For example, Neale~\cite{Neale2012} states how genre labels are not extensive enough to cover the diversity of content within a movie and can be only relevant for the period in which they are used. Altman ~\cite{Altman1984} argues that this is because genres are in a constant process of negotiation and change, where a stable set of semantic givens is developed both through syntactical experimentation and in response to audience or cultural development. We can find thousands of films with identical genre labels but very different inter-textual, semantic content. With this in mind, we propose that genre labels should be considered a weak labelling methodology and to address this classification issue we present a self-supervised approach to finding shared inter-textual information between movies that exist outside of these restrictive genre categories. 

Recent machine learning based genre classification studies, have under-explored the semantic variation that exists between genre labels~\cite{alvarez2019influence,wehrmann2017movie}. Furthermore, efforts have been made to avoid the issues that come with this poorly defined classification problem. In \cite{alvarez2019influence}, the authors show how using a broader range of distribution dates within the movie dataset results in inferior classification when predicted using low-level visual features. We also find that the movie genre classification dataset LMTD-9~\cite{wehrmann2017movie} only features movies from before 1980, which may be in response to the more fluid nature of genre representation in the last thirty years. Lastly, we find that many genre classification papers have avoided multi-label approaches ~\cite{rasheed2005use,huang2012movie,Zhao2016}, simplifying the complex relationships that exist between multiple genres~\cite{Luklow1984}.

Therefore we approach genre classification as a weakly labelled problem, seeking to find similarities between the inter-textual content of movies, within the genre space. To do so, we exploit expert knowledge in the form of semantic embedding `experts' as proposed in \cite{liu2019CollabGate}, obtaining several data modes including scene understanding, image content analysis, motion style detection and audio. Using a contextually gated approach~\cite{Miech2017} enables us to amplify modes that are more useful for multi-label genre-classification, and yields good results for discreet genre labelling. Then inspired by \cite{Luklow1984, Neale2012,Altman1984}, we continue to train the model self-supervised, uniquely leveraging the similarity and differences of sub sequences from within the trailers, to identify inter-textual similarities between the movies for clustering, retrieval, and genre label improvement. This has the effect of expanding genre clusters by their semantic information leading to improved clustering and retrieval . 
\begin{figure*}[tp]
\centering
\includegraphics[width=0.9\textwidth]{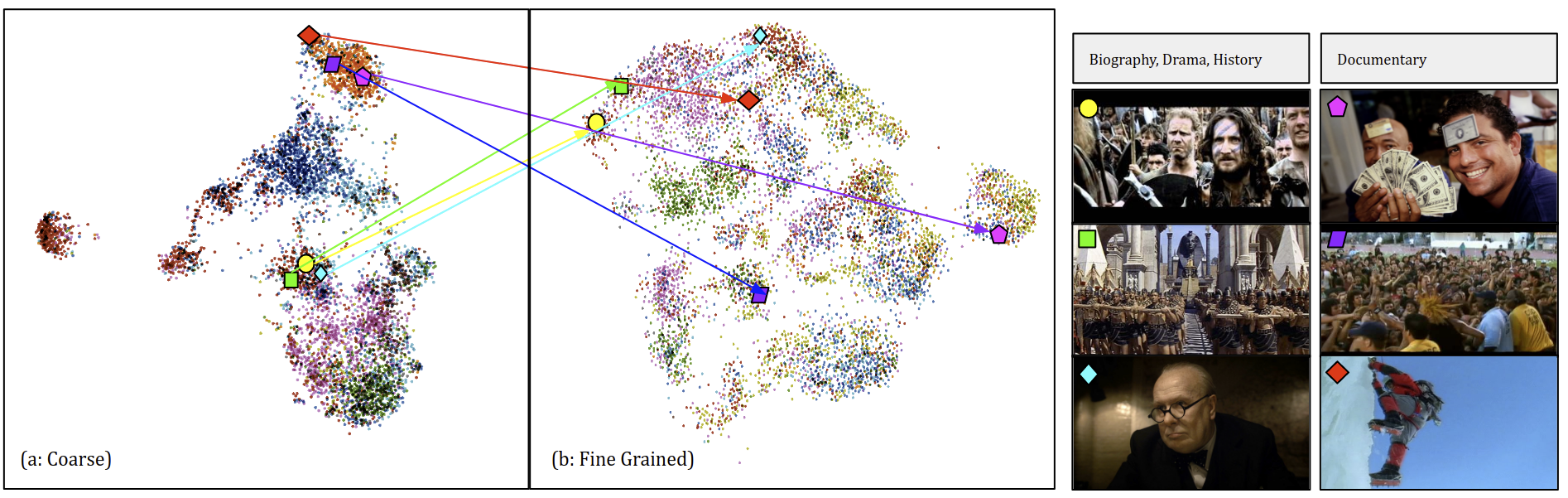}

\caption{Self-supervised Genre clustering via collaborative experts. (a) is a T-SNE plot showing the output of the coarse genre encoder network. Here, trailers that share the same three genres, `Biography', `Drama', and `History' have a high cosine similarity and are well clustered, as is the `Documentary' genre. (b) illustrates the output of our fine grained genre model, where the model has separated the trailers taking into consideration their multi-modal content. In this example, the movie 'Darkest Hour' is pushed further away from `Cleopatra' and `Braveheart' as they share more similar semantic content with other large scale historical action movies. In the second example the three `Documentary' trailers are pushed apart with consideration to their `Music' and `Adventure' inter-textual signatures. }
\label{fig:Motivate}
\end{figure*}

As in other works~\cite{rasheed2005use,huang2012movie,Zhao2016,wehrmann2017movie,shambharkar3DConv}, we use movie trailers as a condensed representation of the content of a movie. Our work is demonstrated in a new large 37 million frame multi-label genre dataset with pre-processed expert embeddings which will be made available for improving research in this field. Our proposed work makes the following contributions: 
\begin{itemize}
\squeezeup\item (i) We introduce MMX-Trailer-20 a new 9K 37,866,450 frame dataset of movie trailers, spanning 120 years of global cinema, and labelled with up to 6 genre labels and 4 pre-computed embeddings per trailer.
\squeezeup
\item (ii) We demonstrate the effectiveness of a multi-modal, collaboratively gated network for multi-label coarse genre classification of up to 20 genres; 
\squeezeup
\item (iii) Enable fine grained semantic clustering of genres via self-supervised learning for retrieval and exploration; 
\squeezeup
\item (iv) The extensive assessment of the performance of the representation on a wide range of genres and trailers; 

\end{itemize}


\section{Related Work}
\squeezeup

\noindent \textbf{Coarse Movie Genre Classification:} Earlier techniques in this field pertain to extracting low-level audiovisual descriptors. Huang(H.Y.) et al.~\cite{huang2007film} used two features - scene transitions and lighting. In contrast, Jain \& Jadon~\cite{jain2009movies} applied a simple neural network with low-level image and audio features. Huang(Y.F.) \& Wang~\cite{huang2012movie} used the SAHS (Self Adaptive Harmony Search) algorithm in selecting features for different movie genres learnt using a Support Vector Machine with good results. Zhou et al.~\cite{zhou2010movie} predicted up to four genres with a BOVW clustering technique. Musical scores have also shown to offer a useful mode for classification as in the work of Austin et al. \cite{austin2010characterization} who predicted genre with spectral analysis using SVMs. More recent work has utilised deep learning and convolutional neural networks for genre classification. Wehrmann \& Barros~\cite{wehrmann2017movie, wehrmann_2016_deep} used convolutions to learn the spatial as well as temporal characteristic-based relationships of the entire movie trailer, studying both audio and video features. Shambharkar et al.~\cite{shambharkarmultimodal} introduced a new video feature as well as three new audio features which proved useful in classifying genre, combining a CNN with audio features to provide promising results. While~\cite{shambharkar3DConv}, employed 3D ConvNets to capture both the spatial as well as the temporal information present in the trailer. The `interestingness' of movies has also been predicted by audiovisual features~\cite{AhmedInterestingness2018}.  Additional features, including text and other metadata, have been combined using simple pooling in more recent work by Bonilla~\cite{cascante2019moviescope} to analyse the complementary nature of different modalities.

\begin{figure*}[thp]
\centering
\includegraphics[width=0.8\textwidth]{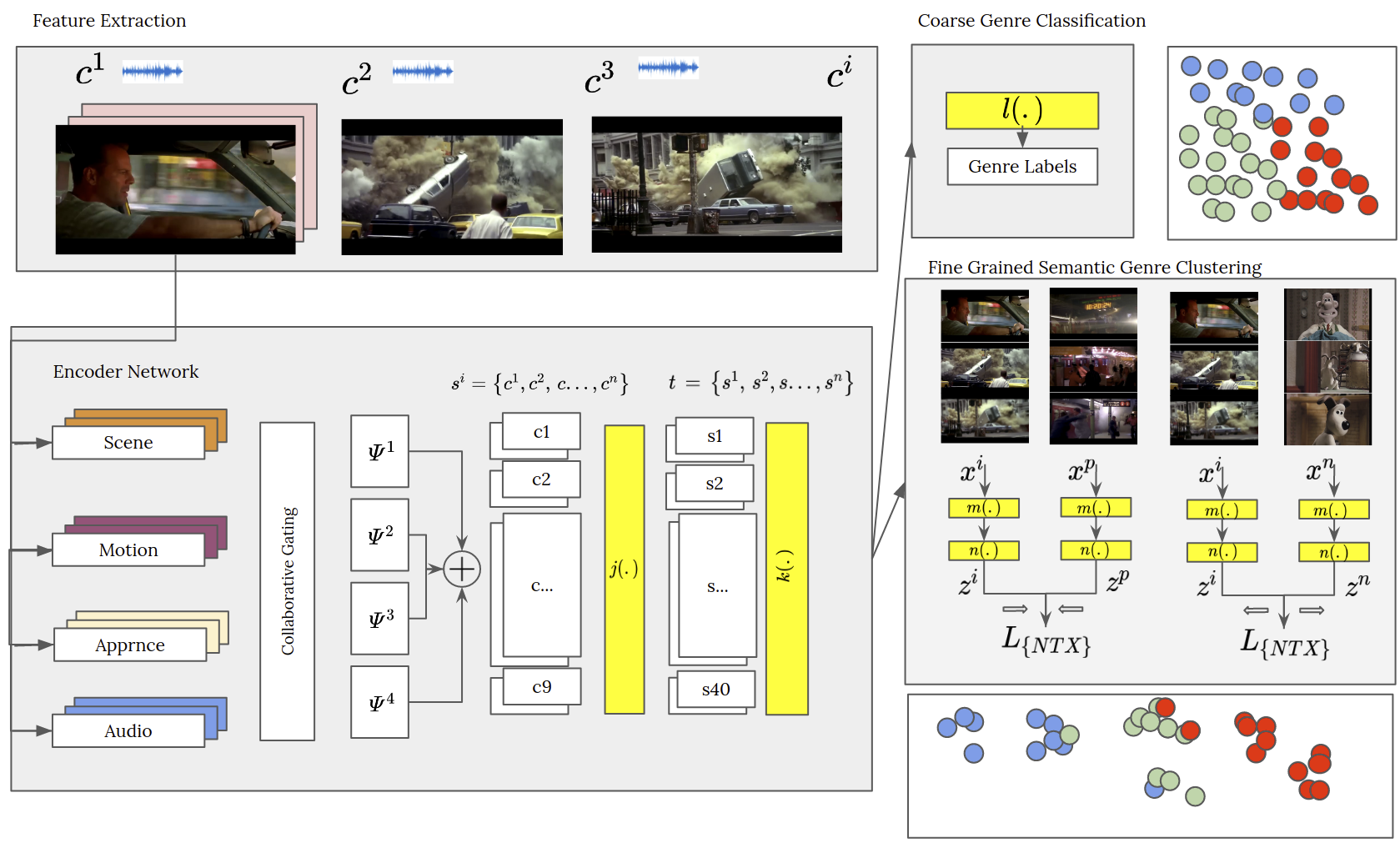}
\caption{An overview of the approach. Where ${c}$ is a clip extracted and ${s}$ is a sequence of 9 clips, constructed from the concatenation of each output from the Collaborative Gating Units. The video sequences are concatenated and passed through the bottleneck MLP ${k(.)}$ which generates the feature embedding vector. Further training for classification encourages this embedding to capture coarse genre information. After training, the whole network is fine-tuned using the self-supervised approach described in Sec \ref{sec:SelfSupervised} to encourage $k(.)$ to highlight similar inter-textual information between samples for fine-grained clustering. Dots here represent broad genres such as Action, Adventure and Sci-Fi. The fine-grained network separates the individual genres, drawing similar films together while retaining some of the broader genre information.}
\label{fig:overiew}
\end{figure*}

\noindent \textbf{Supervised Activity classification:} Given that movies are concatenations of image frames, the field of activity recognition and classification is also relevant. Early seminal works used either 3D (spatial and temporal) convolutions ~\cite{halevy2009unreasonable} or 2D convolutions (spatial)~\cite{he2016identity}, both utilising a mono-mode of appearance information from RGB frames. Simonyan \& Zisserman~\cite{radford2018improving} address the lack of motion features captured by these architectures, proposing two-stream late fusion that learnt distinct features from the Optical Flow and RGB modalities, outperforming single modality approaches. Later architectures have focused on modelling longer temporal structure, through the consensus of predictions over time~\cite{kiros2014unifying,venugopalan2015sequence,yamamoto2003topic} as well as inﬂating CNNs to 3D convolutions~\cite{carreira2017quo_kinetics}, all using the two-stream approach of late fusion of RGB and Flow. Given the temporal nature of video, the computation can be high, therefore recently the focus has been centered on reducing the high computational cost of 3D convolutions~\cite{mithun2018learning,hauptmann2002multi,xu2016msr}, yet still showing improvements when reporting results of two-stream fusion~\cite{xu2016msr}. 

\noindent \textbf{Self Supervision for Activity Recognition:} Given the amount of data video presents, labelling is a challenging and time-consuming activity. Therefore self-supervised methods of learning have been developed. Learning representations from the temporal~\cite{fernando2017self,wei2018learning,miech2017learning} and multi-modal structure of video~\cite{arandjelovic2017look,korbar2018cooperative}, are examples of self-supervised learning, leveraging pre-training on a large corpus of unlabelled videos. Methods exploiting the temporal consistency of video have predicted the order of a sequence of frames~\cite{fernando2017self} or the arrow of time~\cite{wei2018learning}. Alternatively, the correspondence between multiple modalities has been exploited for self-supervision, particularly with audio and RGB~\cite{arandjelovic2017look,korbar2018cooperative,owens2018audio}.


\squeezeup
\section{Methodology}
This section outlines our proposed methodology for both coarse classification and finer grained clustering and retrieval. In Fig.~\ref{fig:overiew}, we present an overview of our approach. Using four pre-trained multi-modal `experts', we extract audio and visual features from the input video. To enable genre classification, a collaborative gating model~\cite{liu2019CollabGate, Miech2017}, learns to emphasise or downplay combinations of these features to minimise a multi-label loss ($L_{BCE}$. This training has the effect of learning the most valuable combination of modes for each label. After we achieve high accuracy for multi-label classification, we encourage the network to develop fine grained semantic clusters through self-supervised training. To achieve this, inspired by the approach of~\cite{chen2020simpleSIMCLR}, we maximise the cosine similarity between sub-sequences within the trailers embedding vectors obtained from the same movie trailer (positive examples) while pushing negative sequence pairs further apart in the feature space.

Given a set of videos $\textbf{v}$, each video is made up of a collection of sequences, $\textbf{s}$, so $\textbf{v} = \{s^1,s^2,...s^m\}$, where there are $m$ sequences in a video. Where each sequence is formed of $n$ clips, giving $\textbf{s} = \{c^1,c^2,...c^n\}$. Ideally, the feature embeddings for all clips should all lie close together as they will have the same class labels, while those from other videos with different labels should lie far apart. The aim of this work is to create a function $\Phi$ that can map a clip $c$ from a video sequence $\textbf{s}$, where $ c \in \textbf{s} \in \textbf{v}$ to a joint feature space $x_i$ that respects the difference between clips. To construct our function $\Phi$, we rely on several pre-trained single modality \emph{experts}, $\{\Psi^1,\Psi^2,...\Psi^E\}$, with $E$ experts and $\Psi^e$ is the $e$'th expert. These operate on the video or audio data and will each project the clip to an individual variable length embedding. Given that the embeddings $\Psi$ are of a variable length, we aggregate the embeddings along their temporal component to form a standard vector size. Any temporal aggregation could be used here, but we use using average pooling for the video based features. While for audio, we implement NetVlad \cite{Arandjelovic2018}, inspired by the vector of locally aggregated descriptors, commonly used in image retrieval. To enable their combination in the following collaborative gating phase, we apply linear projections to transform these task-specific embeddings to a standard dimensionality. 


\subsection{Collaborative Gating Unit}
The Collaborative Gating Unit as proposed in ~\cite{Miech2017} aims to achieve robustness to noise in the features through two mechanisms:
\textbf{(i)} the use of information from a wide range of modalities; \textbf{(ii)} a module that aims to combine these modalities in a manner that is robust to noise.

There is a two-stage process to learn the optimum combination of the expert embeddings for noise robustness: define a single attention vector for the $e$'th expert; then modulate the expert responses with the original data.

To create the $e$'th expert's projection $T^e$, we use the approach first proposed by~\cite{santoro2017simple} for answering virtual questions. The attention vector of an expert projection will consider the potential relationships between all pairs associated with this expert, as defined in eq.~\ref{eq:CollabGatingFeatureCombinations}.
\squeezeup
\begin{equation}
\label{eq:CollabGatingFeatureCombinations}
T^e(\textbf{v})=\textbf{h}( \sum_{\forall E}^{f\ne e}\textbf{g}(\Psi^i(\textbf{v}),\Psi^f(\textbf{v})))
\end{equation}
This creates the projection between expert $e$ and $f$, where $\textbf{g}(.)$ is used to infer the pairwise task relationships while $\textbf{h}(.)$ maps the sum of all pairwise relationships into a single attention vector $T^e$, and $\textbf{v}$ is the set of sequences. Both $\textbf{h}(.)$ and $\textbf{g}(.)$ are defined as multi-layer perceptrons (MLPs). To modulate the result, we take the attention vectors $T=\{T^1(\textbf{v}),T^2(\textbf{v}),...,T^e(\textbf{v})\}$ and perform element wise multiplication (Hadamard product) with the initial expert embedding vector which results in a suppressed or amplified version of the original expert embedding. Each expert embedding is then passed through a Gated Embedding Module (GEM)~\cite{miech2018learning} before being concatenated together into a single fixed length vector for the clip. We capture 9 clip embeddings before concatenating and passing through an MLP to obtain a sequence embedding. These sequence representations are then concatenated together before being passed through a bottleneck layer which learns a compact embedding for the whole trailer.

\squeezeup
\begin{equation}
\label{eq:Hadamard product}
\Psi^{e}(\textbf{v})=\Psi^{e}(\textbf{v})\circ\sigma T^{e}(\textbf{v})
\end{equation}



\subsection{Coarse Grained Genre Classification}
The trailer embedding obtained from the collaborative gating unit $v_i$ can be trained in-conjunction with genre labels to enable classification. Given each trailer can have up to six genre labels, a Binary Cross Entropy Logits Loss is minimised. First the sequence embeddings $x$ are summed over a trailer and then projected via an MLP $\textbf{k}(.)$ to produce a logits embedding. We then proceed to minimise a Binary Cross Entropy Logits Loss until convergence. This loss combines a Sigmoid layer and the Binary Cross Entropy Loss as this is more numerically stable than using a Sigmoid followed by a Binary Cross Entropy Loss.

\begin{align}
\label{eq:3}
\mathcal{L}_{BCE}(u,y)=L_g &=\{l_{1,g},…,l_{N,g}\},  \nonumber      \\
    l_{b,g} =-w_{b,g}[p_g y_{b,g} log\sigma(u_{b,g})&+(1-y_{b,g} log(1-\sigma(u_{b,g}))]
\end{align}
Where $g$ is our genre class label, $b$ is the number of samples in the batch and $p_g$ is the weight or probability of the positive answer for the class $g$, $u$ is the logits and $y$ is the target. With this method it is also possible to perform genre classification on each sequence $s$, by adjusting $\textbf{k}(.)$ so that $x$ becomes the logits embedding. While the gated encoder accuracy is degraded slightly by this technique (as outlined later in the ablation studies, see Tbl.~\ref{tab:ExpertAblation}) it is possible to identify different sub genres at a sequence level. For example one could identify specific `Adventure' sequences in a movie that only has the genre label of `Action'. 

We show in the results section later how collaborative gating is effective at improving the prediction task of user-defined labels in a fully supervised manner. By analysing data in the bottleneck vector, we can see how the model is able to capture important semantic information for genre clustering. However, we can also see that movies with similar genre labels are grouped, even if the style or content of the movie varies. To create a more granular representation of the genre space, we continue to allow the model to train self-supervised. 


\subsection{Fine Grained Semantic Genre Clustering}
As discussed in the introduction, discreet genre labels are restrictive and only offer a broad representation of the content of a video. We aim to find finer grained semantic content by identifying similarities in the sound, locations, objects, and motion within the videos. To achieve this, we extend the pre-trained coarse genre classification model with a self supervised contrastive learning strategy using a normalised temperature-scaled cross-entropy loss (NT-Xent) as proposed by Chen \cite{chen2020simpleSIMCLR}, $\mathcal{L_{NTX}}$. In \cite{chen2020simpleSIMCLR}, image augmentations are used as comparative features to fine tune the embedding layer of their classification network. The goal is to encourage greater cosine similarity between embeddings obtained from the same image, while forcing the negative pairs apart. We uniquely extend this method to video, by splitting each movie trailer into two equal lengths of sequences, and using the embeddings of these sequences as the representation pairs $x$.
\begin{equation}
\label{eq:cross-entropyLoss}
\mathcal{L_{NTX}}(x) = -\log\frac{exp(\mathrm{sim}(\textbf{m}(x_i), \textbf{m}(x_p)/\tau)}{\sum_{k\neq j}^{2N} \mathbb{1}_{n\neq p}exp(\mathrm{sim}(\textbf{m}(x_i),\textbf{m}(x_n))/\tau}
\end{equation}

Here  $x_i$, $x_p$ and $x_n$ are the feature representations and $\textbf{m(.)}$ represents a projection head encoder formed from MLPs, $\tau>0$ is a temperature parameter set at 0.5 and $\mathrm{sim}$ is the cosine similarity metric. $x_i$ and $x_p$ are two embedding vectors obtained from the same video as described above, while $x_n$ is an embedding vector from another video. Here, the $\mathcal{L}_{NCE}$ loss will enforce $x_i$ closer in cosine similarity to $x_p$ but further from $x_n$. This process is illustrated in the overview Fig.~\ref{fig:overiew}.

Pairing each embedding vector from the video $\textbf{s}$ with all other embedding vectors from other videos, will maximise the number of negatives. As a result, for each video, we get $2 \times(N-1)$ negative pairs — where $N$ is the number of videos in the dataset. Therefore in training we mini-batch sequences, which comprises of $2 \times(N-1)$ sequences $B=\{b_1,b_2,...,b_{2 \times(N-1)}\}$. The overall contrastive loss is computed as shown in the equation below

\begin{equation}
\label{eq:cross-entropyLossSum}
\mathcal{L}_{CON}(B) = \sum_{i=1}^{2 \times(N-1)} \mathcal{L}(i)
\end{equation}
After training, the  MLP projection head ($\textbf{m}(.)$) is removed and we use the bottleneck layer of the collaborative gating model as a pre-trained embedding projection network. The fine-tuning using a contrastive loss encourages the bottleneck layer to retain some coarse genre information while finding similar inter-textual elements in other trailers. This leads to a more diverse clustering of samples, identifying sub-label information within the original label clusters.

\squeezeup
\section{Results and Discussion}
This section introduces the MMX-Trailer-20 dataset in Sec.~\ref{sec:dataset}, followed by specific architecture and implementation details in Sec.~\ref{sec:Implementation}. Key results for coarse genre classification are presented in Sec.~\ref{sec:baseline} with an ablation study of the method’s components and comparison against other baselines. Fine grained semantic clustering of the dataset is then presented in Sec.~\ref{sec:SelfSupervised} (on self-supervised retrieval).

\subsection{MMX-Trailer-20: Multi-Model eXperts Trailer Dataset}
\label{sec:dataset}
There are several datasets upon which previous works test. However, to capture the scale and variability of a dataset is challenging, especially in terms of diversity of genre, size of dataset and year of distribution. Tbl.~\ref{tab:DatasetComparsion} shows the comparison in size and labelling between recent works in genre classification.
\begin{table}[htbp]

\caption{The details of other Movie genre datasets}
\centering
\begin{adjustbox}{width=0.48\textwidth}
\begin{tabular}{l|cccccc}
\toprule
\multirow{2}{*}{Dataset}            & Video  & Number &\multirow{2}{*}{Frames}& Label           &Num. &Genre/ \\ 
                                    & Source & Trailers&          &  Source         &Genres & Trailer    \\
\midrule

Rasheed~\cite{rasheed2005use}       &Apple   & 101     &  -         &   -             & 4      &  1           \\
Huang~\cite{huang2012movie}        &Apple   & 223     &  -         &   IMDb          &  7     &   1           \\
Zhou~\cite{Zhao2016}                &IMDb+Apple& 1239 & ~4.5M & IMDb            & 4      & 3          \\
LMTD-9~\cite{wehrmann2017movie}     & Apple  & 4000    & ~12M&  IMDb           & 9      &  3           \\
Moviescope~\cite{cascante2019moviescope}& IMDb& 5000&  ~20M    & IMDb     & 13      & 3       \\ 
\midrule
MMX-Trailer-20                        &Apple+YT    & 8803 &  ~37M  &  IMDb           & 20     & 6        \\\bottomrule

\end{tabular}
\end{adjustbox}
\label{tab:DatasetComparsion}
\squeezeup
\end{table}

\begin{figure*}
     \centering
     \begin{subfigure}[b]{0.245\textwidth}
         \centering
         \includegraphics[width=\textwidth]{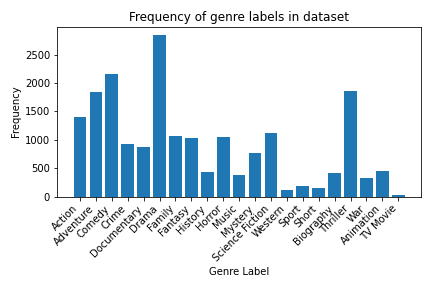}
         \caption{Freq of genre in dataset}
         \label{fig:GenreinDataset}
     \end{subfigure}
     \begin{subfigure}[b]{0.245\textwidth}
         \centering
         \includegraphics[width=\textwidth]{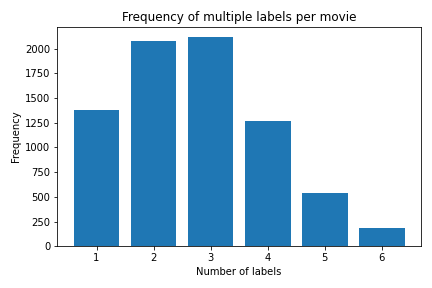}
         \caption{Num of labels/trailer}
         \label{fig:NumLabelsPerTrailer}
     \end{subfigure}
     \begin{subfigure}[b]{0.245\textwidth}
         \centering
         \includegraphics[width=\textwidth]{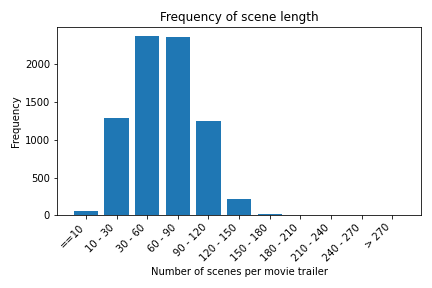}
         \caption{Scene length}
         \label{fig:SceneLength}
     \end{subfigure}
     \begin{subfigure}[b]{0.245\textwidth}
         \centering
         \includegraphics[width=\textwidth]{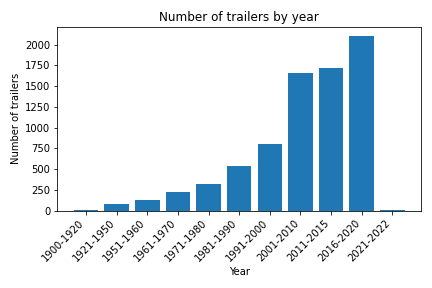}
         \caption{Year of distribution}
         \label{fig:MovieYears}
     \end{subfigure}
        \caption{MMX-Trailer-20 Dataset statistics (best zoomed in).}
        \label{fig:datasetStats}
\squeezeup
\end{figure*}

As shown, most datasets are small with limited numbers of genre labels, both in terms of variability and the number assigned to a single trailer. Moviescope \cite{cascante2019moviescope} is the closest to the proposed dataset, with 3 genre labels and 5000 trailers.  However, we increase the number of trailers and labels per trailer, while increasing the number of frames available by order of magnitude. Furthermore, we will make available all pre-computed expert embeddings for every scene. The collection totals 8803 movie trailers drawn from Apple Trailers and YouTube, comprising 37,866,450 individual frames of video. The statics of the dataset can be seen in Fig.~\ref{fig:datasetStats}, for example, a wide range of genres exist, and each trailer is labelled with on average at least 3 genres, while the year of the trailers is diverse from 1930s to the present day.

\begin{figure}[htp]
\centering
\includegraphics[width=0.45\textwidth]{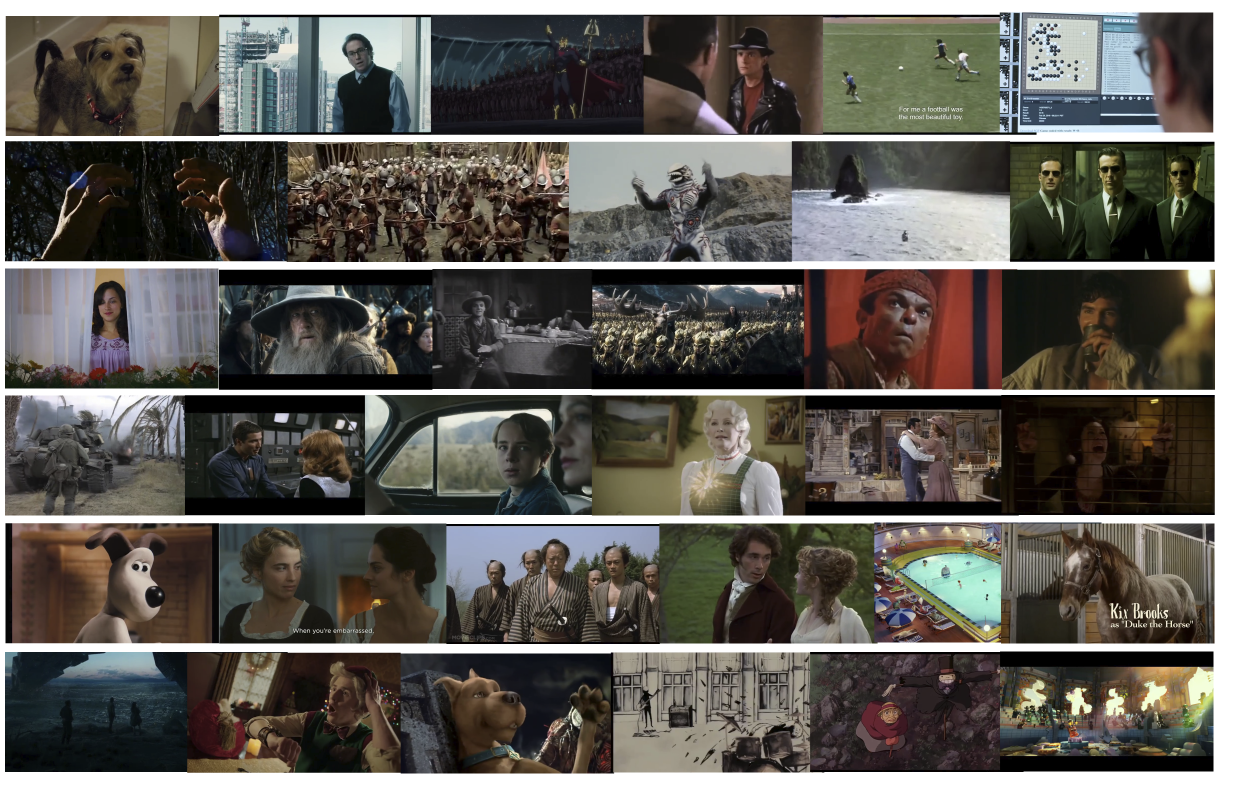}
\caption{Examples of the diversity of trailers within the MMX-Trailer-20 dataset}
\label{fig:DatasetEx}
\end{figure}

Every trailer is a compact encapsulation of the full movie through a short 2 to 3 minute video, and we can collect a weak proxy of genre classification by matching the trailer to its user generated entry on the website \url{imdb.com}. On IMDb, users can select up to six genre labels for each trailer. The dataset has 20 genres - Action, Adventure, Animation, Comedy, Crime, Documentary, Drama, Family, Fantasy, History, Horror, Music, Mystery, Science-Fiction, Western, Sport, Short, Biography, Thriller and War. Qualitative examples illustrating the variety of the dataset are shown in Fig.\ref{fig:DatasetEx}. 

\noindent\textbf{Processing of the Dataset:} The dataset is pre-processed, with scene detection performed using the PyScene Detect~\cite{scenedetectionoverview}, extracting individual clips from each trailer.  While, we remove the first and last frames to mitigate poor scene detection. Audio is extracted as a mono 16bit Wav file at 16khz using FFmpeg. To compute motion frames, we use Dual $TVL1$ Optical Flow as introduced in~\cite{zach2007duality} and outlined in the implementation of~\cite{TVL1Implementation}, before passing the optical flow images via the motion expert encoder. Extracted frames are also passed to the scene and appearance encoders. We partition the dataset into 6047 trailers for training, 754 for validation and 754 for testing totalling 7555 trailers. The number of trailers used for evaluation is 1248 less than the dataset as we exclude trailers that have less than ten clips. This is done so we can maintain constant batch sizes at a minimum sequence size.



\subsection{ Implementation Details}
\label{sec:Implementation}
\noindent \textbf{Expert Features:} 

To capture the rich content of the trailer, we draw on several existing powerful representations that are present in movie trailers, \emph{Appearance, Audio, Scene and Motion}. The \textbf{\emph{Appearance}} feature is extracted using an SENet154 model~\cite{hu2018squeeze}, pre-trained on ImageNet~\cite{imagenet_cvpr09} for the task of image classification, creating a $1\times2048$ embedding. The \textbf{\emph{Scene}} feature is computed on a per frame basis from a ResNet-18 model~\cite{he2016deep} pre-trained on the Places365~\cite{zhou2017places} dataset, returning a $1\times1024$ embedding. The \textbf{\emph{Motion}} of the clip is encoded via the I3D inception model~\cite{carreira2017quo_kinetics} and a 34-layer R(2+1)D model~\cite{tran2018closer} trained on the kinetics-600 dataset~\cite{carreira2017quo_kinetics}, producing a $1\times1024$ embedding. The \textbf{\emph{Audio}} embeddings are obtained with a VGG style model, trained for audio classification on the YouTube-8m dataset~\cite{Abu-El-Haija2016} resulting in a $1\times128$ embedding. To aggregate the features extracted on a frame-wise basis, for appearance, scene and motion embeddings, we average frame-level features along the temporal dimension to produce a single feature vector per clip per feature. For audio, we aggregate the features using a vector of locally aggregated descriptors as outlined in \cite{arandjelovic2017look,Arandjelovic2018}. We then average each expert feature to the same dimension of $1\times768$ before passing to the collaborative gating unit described above. 

\noindent \textbf{Training details:} 

We implement our model using the PyTorch library, and hyper-parameters were identified using coarse to fine grid search. For supervised coarse genre classification, the Binary Loss is reduced over 200 epochs using the Adam Optimiser~\cite{Kingma2015} with AMSGrad~\cite{Reddi2018}, and with an initial learning rate of 0.00003, and a batch size of 32 samples. In the self-supervised training, we adjust the learning rate to 0.0001. We pass ten epochs of the samples through the encoder before reducing the learning rate using cosine annealing as proposed in \cite{chen2020simpleSIMCLR}. We continue to fine-tune the network for 50 epochs. Once the semantic encoder has been fine-tuned, we remove the projection head network and then use the output of the bottleneck layer at run time. 

\begin{table*}[htbp]
\centering
\caption{Coarse genre classification of the MMX-Trailer-20 dataset. Across differing expert features and combinations methods (note $(\overline{PRC})= AU(\overline{PRC})_w$) }
\begin{adjustbox}{width=0.98\textwidth}
\begin{tabular}{l|cccccccccccccccccccc|cccc}
Model                                 &Actn &Advnt&Animtn&Bio &Cmdy&Crme&Doc &Drma&Famly&Fntsy&Hstry&Hrror&Mystry&Music&SciFi&Wstrn&Sprt &Shrt&Thrll&War&$F1_w$  &$(\overline{PRC})$&$P_w$     & $R_w$\\ 
\midrule
Support                               & 130 &197  &46    &13  &224 &102 &87  &267 &117  &115  &44   & 104 &41   &86   &107  &181  &30   &45  &12   &21  & -     & -                & -        & -\\
\midrule
Random                                &0.29 &0.41 &0.11  &0.03&0.46&0.24&0.21&0.52&0.27 &0.26 &0.11 & 0.24&0.1   &0.2  &0.25 &0.39 &0.08 &0.11&0.03 &0.05&0.318  &    	0.134	  &0.19	     &1   \\
Scene~\cite{he2016deep}               &0.43&0.55  &0.74  &0   &0.49&0.38&0.63&0.55&0.51 &0.28 &0.24 &0.42 &0.3	 &0.28&0.41  &0.51 &0.22 &0.19&0.11 &0.33&0.434  &	0.489	         &0.437     &	0.48    \\        
Audio~\cite{Abu-El-Haija2016}         &0.47&0.51  &0.40  &0.10&0.61&0.38&0.58&0.55&0.51&0.37 &0.11  &0.34 &0.39  &0.30&0.35  &0.55 &0.16 &0.15&0.13 &0.12&0.454  &	0.449	         &  0.400   &0.537    \\
Motion~\cite{carreira2017quo_kinetics}&0.5 &0.59  &0.74	 &0   &0.62&0.33&0.63&0.56&0.55 &0.36 &0.2  &0.38&0.45  &0.24&0.37  &0.57 &0.23 &0.14&0.10 &0.13&0.463  &	0.487	         &0.448	    &0.494   \\
Image~\cite{hu2018squeeze}            &0.48&0.63  &0.79  &0.12&0.65&0.41&0.60&0.59&0.55&0.42 &0.25  &0.47&0.42  &0.29&0.50  &0.54 &0.34 &0.19&0.12 &0.31&0.516  &	0.554            &0.493     &0.572    \\
Image + Audio                    &0.52&0.63&0.78&\textbf{0.15}&0.65&0.42&0.68&0.6  &0.63&0.46  &0.25&0.50 & 0.51&0.34&0.49  &	0.59 &0.38 &\textbf{0.28}&0.12&0.42&0.544  &0.558	 &0.476	    & 0.65   \\
Image + Motion                           &0.59&0.64 &0.78 &0 &0.59 &0.39 &0.66 &0.6 &0.6 &0.5 &0.29 &0.54 &0.53 &0.25 &\textbf{0.52} &0.57 &0.4 &0.2 &0.24                 &0.12&0.535  &0.553    &0.511     &0.583 \\
Image + Scene                &0.52 &0.61 &0.80  &0.12 &0.61&0.37&0.65&\textbf{0.62}&0.58&0.49  &0.15&0.51 &0.49 &0.37	&0.48 &0.56	 &\textbf{0.43}&0.26&0.12&0.46&0.531  &0.539     &	0.490	&0.600 \\
Naive Concat                             &0.56&0.61  &0.64  &0.09&0.64&0.35&0.69&0.60&0.58&0.39&0.19&0.49 &0.45 &0.21    &0.48 &0.6   &0.39 &0.28 &0.27&	0.41&0.525 &0.497             &0.522    &0.551\\
MMX-Trailer-20 &\textbf{0.62}&\textbf{0.69}&\textbf{0.71}&0.11&\textbf{0.71}&\textbf{0.53}&\textbf{0.73}&\textbf{0.62}&\textbf{0.64}&\textbf{0.51}&\textbf{0.34}&\textbf{0.56}     &\textbf{0.60}&\textbf{0.45} &0.50&\textbf{0.64}&0.30&0.11&\textbf{0.13}&\textbf{0.55}&\textbf{0.597}&\textbf{0.583}&\textbf{0.554}&\textbf{0.697} \\ 
\bottomrule

\end{tabular}
\end{adjustbox}
\label{tab:ExpertAblation}
\end{table*}

\noindent \textbf{Evaluation Metrics:} 
We use the standard retrieval metrics as proposed in prior work~\cite{dong2016word2visualvec,Miech2017,mithun2018learning}. Given the variance of the frequency of occurrence of the genre labels in the dataset we employ the following metrics designed to cope with unbalanced data: $\overline{AU(PRC)}$ (micro average),
$AU(\overline{PRC})$ (macro average), and $AU(\overline{PRC})_w$
(weighted average). Each measure emphasises different aspects regarding the method’s performance. The $\overline{AU(PRC)}$ measure averages the areas of all labels, which causes less-frequent classes to have more influence in the results. To ensure that we are performing well across all categories, even for those that have less training samples or are more difficult to predict. $AU(\overline{PRC})$ uses all labels globally, which makes high-frequency classes have greater influence in the results, ensuring that we obtain overall good results across all samples in the dataset. Finally, $AU(\overline{PRC})_w$ is calculated by averaging the area under precision-recall curve per genre, weighting instances according to the class frequencies, allowing each sample to be measured independently from the whole set, and then gives us an averaged score. We also show weighted Precision  ($P_w$) and weighted Recall ($R_w$) and weighted F1-Score ($F1_w$), for all metrics higher is better.

\squeezeup
\subsection{Coarse Grained Genre Classification Results}
\label{sec:baseline}

We analyse the performance of our approach both quantitatively and qualitatively for both classification and self-supervised retrieval on \emph{MMX-Trailer-20}, the trailer dataset. Tbl.~\ref{tab:ExpertAblation} illustrates the quantitative performance of the coarse genre prediction model \emph{MMX-Trailer-20} intra genre, and the global metrics. The table also shows the random performance, which will vary according to the frequency of the genre in the dataset. and also identity \textbf{random} performance for the metrics. 
Also, it explores the influence of each of the individual experts on the coarse genre classification task. Individual experts are passed directly through to the first MLP, while pairs are collaboratively gated as outlined in Sec. \ref{sec:Implementation}. From these results, we can see that the image expert is most valuable for genre classification and becomes more effective when combined with motion. Using collaborative gating yields a 10\% increase in basic fusion through concatenation. Audio and Scene are the weakest experts for the classification task, which may be due to features that are not genre specific such as dialogue, and external environments. 
We find all Visual experts perform best on Animation, most likely due to its unique style from the other trailers while the Audio expert performs better on Comedy and Sport. To identify the importance of the collaborative gating units, we compute a naive concatenation of the feature embeddings from the experts passed through an MLP layer (\textbf{Naive Concat}). This is shown to have a 10 point reduction compared to using the gating to aggregate the features, illustrating the importance of the learnt collaborative gating framework.

To attempt a comparison to other approaches, Tbl.~\ref{tab:PriorBaselines} shows the best performance of other approaches on different datasets.  Our model, \emph{MMX-Trailer-20} uses up to 6 genre labels per sample from a total of 20 genres, double most other approaches and will affect the random baseline which is nearly half that of the 9 genre datasets. To contextualise our method with others we compare previous approaches including video low level features (\textbf{VLLF})~\cite{rasheed2005use}, audio-visual features (\textbf{AV})~\cite{huang2012movie, cascante2019moviescope}, audio-visual features with convolutions over time \textbf{CTT-MMC}~\cite{wehrmann2017movie}, and an \textbf{LSTM} model that uses visual feature data in a standard sequence analysis approach as implemented for comparison in~\cite{wehrmann2017movie}. From the results in Tbl.~\ref{tab:PriorBaselines} we show that our model performs better than low-level features as well as the LSTM model. We do not improve performance on other audiovisual approaches which fine-tune pre-trained networks in an end to end manner \cite{wehrmann2017movie, cascante2019moviescope} which use a far smaller subset of genre and labels in their older datasets. 

\begin{table}[htbp]

\centering
\caption{Comparison of our proposed approach with the other methods for genre classification.}
\begin{adjustbox}{width=0.48\textwidth}
\begin{tabular}{l|cc|ccc}
\toprule

Method                                   & no genres & no labels &$\overline{AU(PRC)}$&$AU(\overline{PRC})$&$AU(\overline{PRC})_w$    \\
\midrule
Random 9 Class                          &   9       & 3         &  0.206        &  0.204        &  0.294            \\    
Random 20 Class                         &   20      & 6         & 0.134         & 0.130         &0.208 \\   
VLLF~\cite{rasheed2005use}              &   9       & 3         & 0.278         &0.476          & 0.386             \\
AV~\cite{huang2012movie}                &   9       & 3         & 0.455         &0.599          &0.567              \\
LSTM~\cite{wehrmann2017movie}           &   9       & 3         & 0.520         &0.640          &0.590              \\
CTT-MMC~\cite{wehrmann2017movie}        &   9       & 3         & 0.646         &0.742          &0.724              \\ 
Moviescope~\cite{cascante2019moviescope}& 13        & 3         & 0.703          &0.615           & -      \\
\midrule
Proposed  MMX-Trailer-20                & 20        &  6        &0.456          &0.589          & 0.583     \\
\bottomrule

\end{tabular}
\end{adjustbox}
\label{tab:PriorBaselines}
\end{table}



\subsection{Fined Grained Genre Exploration}
\label{sec:SelfSupervised}

\begin{figure}[htp]
\centering
\includegraphics[width=0.5\textwidth]{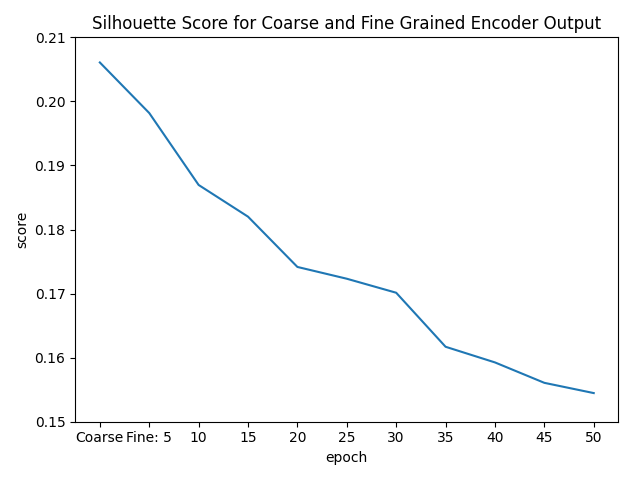}
\caption{Silhouette score~\cite{SilhouetteScore} of the coarse encoder output and then the following 50 epochs of fine-tuning to develop the fine grained model. The decreasing score shows that the tight discrete genre clusters are separating and overlapping more as we fine-tune the network.}
\label{fig:silhouette}
\squeezeup
\end{figure}

While the coarse genre classification is interesting, in general, discreet labels are limited in providing a full understanding of complex trailers. It is challenging to quantify this fully. However, we evaluate the effectiveness of the self-supervised fine grain genre learning by comparing the cosine similarity between embedding trailer vectors before and after being processed by the fine-grained self-supervised network. This is visualised in a T-SNE plot in Fig.~\ref{fig:Motivate}, where the colours indicate the primary genre. Fig.~\ref{fig:Motivate}(a) shows the learnt embedding for the coarse genre classification, where tight genre clusters are formed. Fig.~\ref{fig:Motivate}(b) is after the self supervised training of the model, where we can see how the clusters have broken up into an overlapping distribution as genres are separated depending on the multi-modal content present in the trailer. We have identified three trailers (Cleopatra, Braveheart, and Darkest Hour) which share the triple genre classification of \emph{Drama, Biography, History} (identified by the three shapes). These are correctly labelled by the coarse genre encoder, have a high cosine similarity and in Fig.~\ref{fig:Motivate}(a), are spatially close in the coarse genre T-SNE plot. In Fig.~\ref{fig:Motivate}(b), after self-supervised training, the trailer embeddings have higher cosine similarity to other genres. For example, we find that Cleopatra is drawn closer to Adventure films featuring deserts and orchestral scores (Lawrence of Arabia is one example). Braveheart shares a high cosine similarity with medieval and mythological trailers featuring large scale battles, while Darkest Hour moves towards a cluster featuring historical thrillers such as 'The Imitation Game'. This effect is quantified in Fig.~\ref{fig:silhouette}, which shows the results of the silhouette score~\cite{SilhouetteScore} of the embedding space during the fine grained training phase. The decreasing score shows that the tight but restrictive genre classification of the coarse model are being broken and genre overlapping is occurring as the training continues.

\begin{figure}[htp]
\centering
\includegraphics[width=0.5\textwidth]{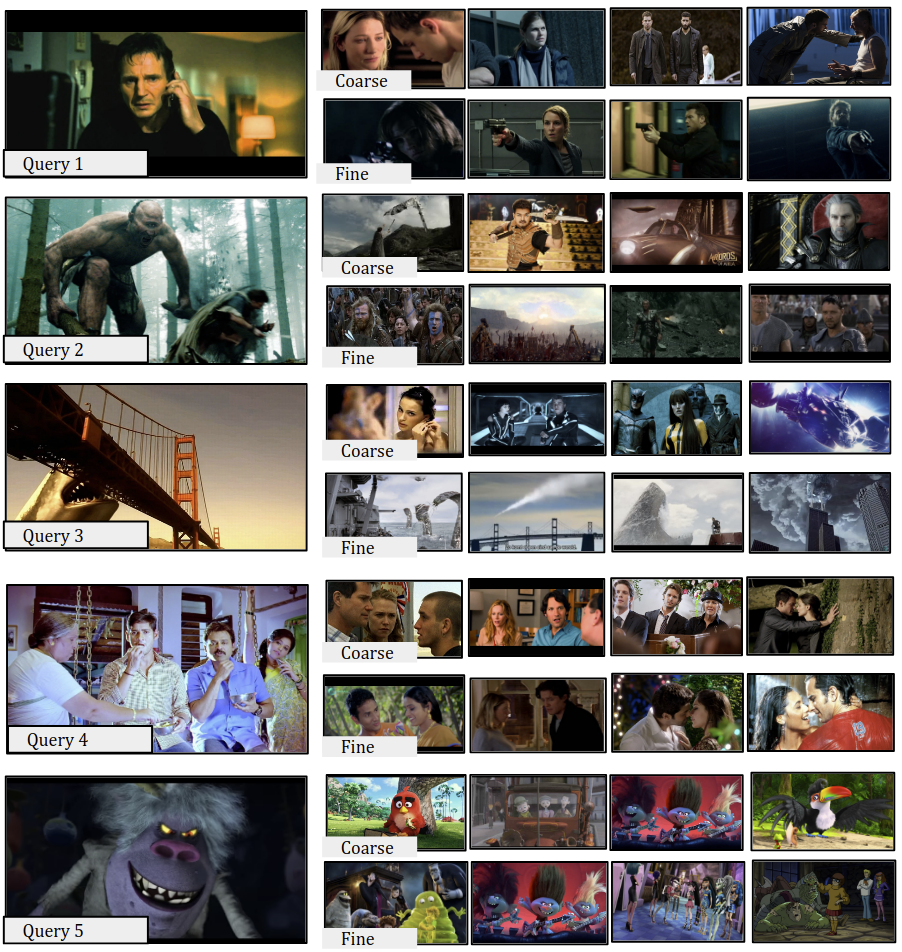}
\caption{Representative retrieval results for 5 queries on the Self-Supervised feature space. We find that the model is able to retrieve trailers with greater contextual awareness than is possible from just classification or Unsupervised training.}
\label{fig:img_retrieval}
\squeezeup
\end{figure}

We can also show illustrative retrieval results. In Fig.~\ref{fig:img_retrieval} we provide five further retrieval examples. The first example Trolls (2016) not only retrieves its sequel, Trolls World Tour (2020) but also retrieves animated family movies featuring monsters. Retrieval four, Mega Shark Vs Giant Octopus (2009), has highest cosine similarity to other sea-monster and environmental disaster movies such as Bemuda Tenticles (2014) and The Meg (2018). In retrieval 5, Seethamma Vakitlo Sirimalle Chettu (2013), we discover that the model clusters other Telugu Language Films demonstrating the model's ability to identify a cultural context within genre clusters. However, we also find Bridget Jones' Diary (2001) within this cluster, suggesting that the cluster has not been completely isolated from other romantic comedies. These results demonstrate greater depth and nuance when compared to retrieval using the coarse genre classifier. It is important to note that in the returned results the genre has not been determined using low-level pixel information. Instead, they uncover an interesting semantic and inter-textual relationship between style, sound, and content of different movie genres and production techniques.

\subsection{Augmentation of Genre Labels}
It is also possible to use the self-supervised network to augment and improve the overall labelling of the original movie trailers, as shown in Tbl.~\ref{tab:Improvedlabels}. To test this, we compared genre labels produced by IMDb to find mislabelled examples in the IMDb dataset. We then asked the network to produce labels based on a sigmoid threshold of 0.30. The model was able to extend and match the IMDb labels and also offered additional labels which are logical concerning the trailer. 

\begin{table}[htbp]
\centering
\caption{Example of the improved genre labelling of movies. Original genre is the label sourced from IMDb and Predicted Genres is the result of proposed model. Blue indicates additional predicted labels.}
\begin{adjustbox}{width=0.48\textwidth}
\begin{tabular}{c|c|c}
Movie                    &IMDb labelled Genres        &Predicted Genres         \\ 
\midrule
101 Dalmatians           &\multirow{2}{*}{Action, Family}&\textcolor{blue}{Adventure},\textcolor{blue}{Comedy},        \\
       II                &                          &Family,\textcolor{blue}{Fantasy},\textcolor{blue}{Animation} \\
\midrule
300                      &\multirow{2}{*}{Action, Drama}&Action, \textcolor{blue}{Adventure},                    \\
Rise of an Empire        &                          &Drama, \textcolor{blue}{Fantasy}                    \\
\midrule
Alien-                   &Horror, Sci-Fi,           &\textcolor{blue}{Action},\textcolor{blue}{Adventure}\\
Covenant                 & Thriller                 &Horror, Sci-Fi                                         \\
\midrule        
\multirow{2}{*}{Company Of Heroes}&\multirow{2}{*}{Drama,War}&War, \textcolor{blue}{History},              \\
                         &                          &Drama, \textcolor{blue}{Action}              \\
\midrule
Independence Day         &Action,                   &Action, Sci-Fi,   \\
Resurgence               &Sci-Fi           &\textcolor{blue}{Adventure}, \textcolor{blue}{Thriller}       \\ 
\midrule
Laws of                  &\multirow{2}{*}{Comedy}   &Comedy, \textcolor{blue}{Crime},          \\
Attraction               &                          &       \textcolor{blue}{Drama}            \\
\midrule
Leprechaun               &Comedy, Fantasy,          &\textcolor{blue}{Adventure}, Comedy,   \\
Returns                  & Horror                   & Fantasy, Horror                        \\
\midrule
\multirow{2}{*}{Santa Paws 2}&\multirow{2}{*}{Family}&Family, \textcolor{blue}{Comedy},            \\
                         &                          &\textcolor{blue}{Adventure}, \textcolor{blue}{Fantasy}        \\
\midrule
The Hobbit: The Battle   &Action,                   &Adventure, \textcolor{blue}{Fantasy},        \\
of the 5 Armies          &Adventure,                & Action, \textcolor{blue}{Sci-Fi}  \\
\midrule
The Land Before Time     &Family                    &Adventure,\textcolor{blue}{Fantasy}   \\
VIII                     &  Adventure               &Family, \textcolor{blue}{Animation}                      \\
                         &fantasy                   &   Fantasy  \\
\bottomrule
\end{tabular}
\end{adjustbox}
\label{tab:Improvedlabels}
\end{table}


\squeezeup
\squeezeup

\section{Effect of sequence length}

\begin{figure}[htp]
\centering
\includegraphics[width=0.4\textwidth]{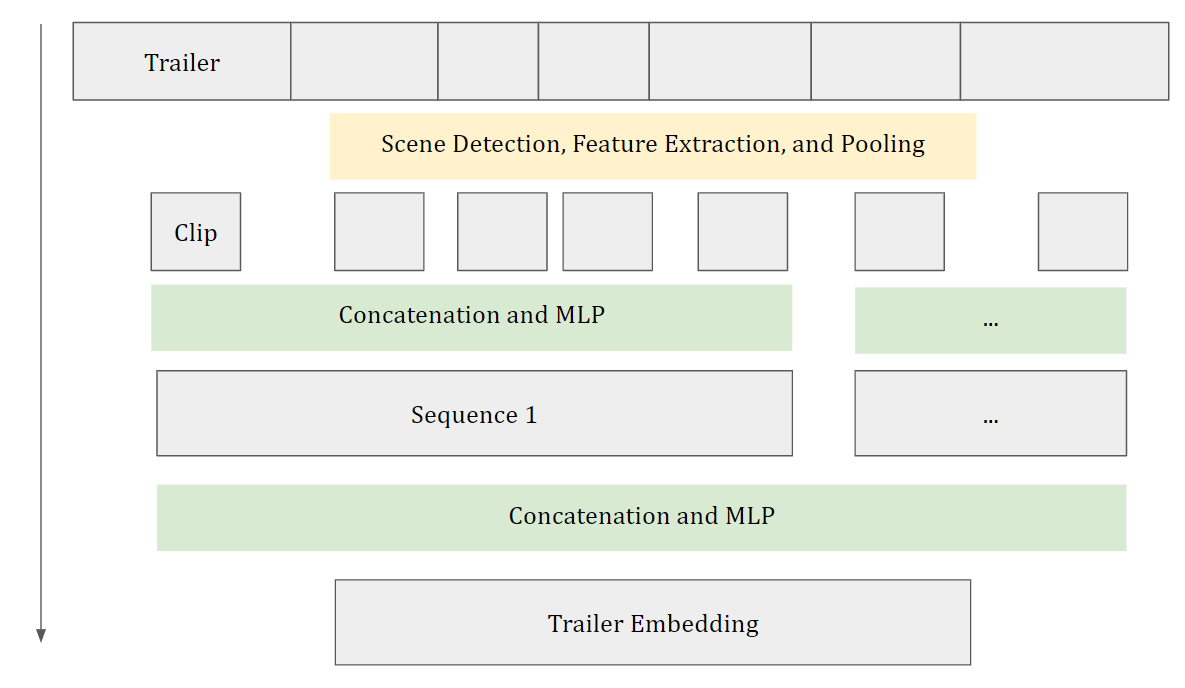}
\caption{Splitting of trailers into clips and sequences. Here a sequence length of five is shown but a length of nine was used in the implementation.}
\label{fig:splitting}
\squeezeup
\end{figure}

In \cite{wehrmann_2016_deep} it's shown how capturing temporal information using both 3D convolutions and LSTM's can help with the genre classification task. To see if temporal information could be retained through concatenation of scenes and sequences, we experimented with a number of scene length variations. Fig \ref{fig:splitting} shows how we extracted clips and constructed sequences. First we used the scene detection method as outlined in the paper to extract individual clips before performing feature extraction and pooling to create equal 1 x 768 embeddings for every mode. Following collaborative gating, each attention vector is then concatenated into a sequence embedding. The concatenated sequences are then passed to the MLP before being concatenated with all other sequences to form a feature embedding for the whole trailer. In the case of self-supervised learning we select random sequences from the same trailer for comparison. 

\begin{figure*}[htp]
\centering
\includegraphics[width=1.0\textwidth]{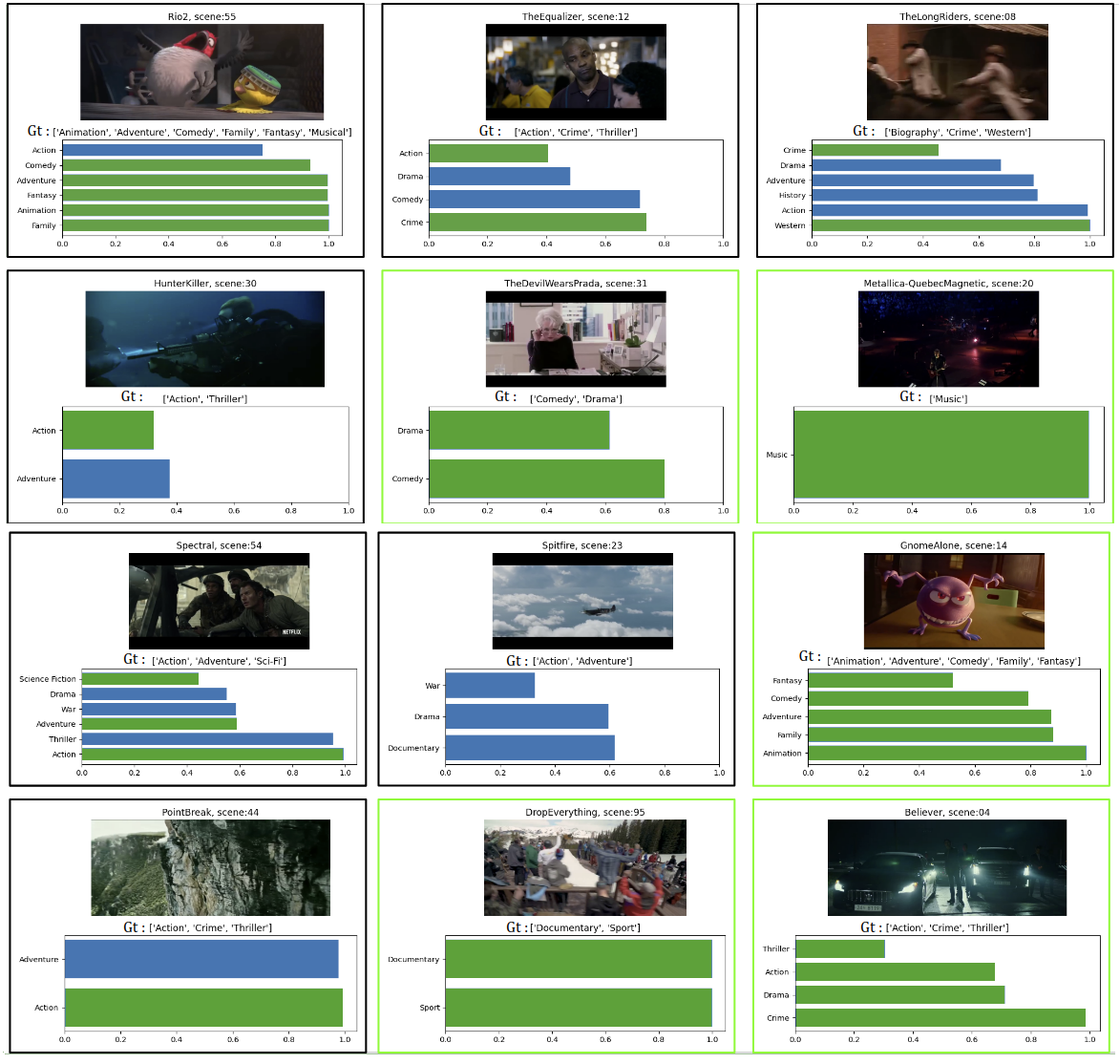}
\caption{Representative results for multi-label classification on single scenes. Green represents results where the model predicts the correct number and correct labels for the scene. Black indicates where the model suggests alternative genres for the scene.  We found that the model was able to make adequate guesses at the individual scene genre when not presented in context with the whole trailer. For example scene 44 from Point Break (bottom left) makes the prediction `Adventure, Action' which would be a good prediction given the image, camera motion, and audio of the scene.}
\label{fig:scene_preds}
\squeezeup
\end{figure*}
\begin{figure*}[htp]
\centering
\includegraphics[width=1.0\textwidth]{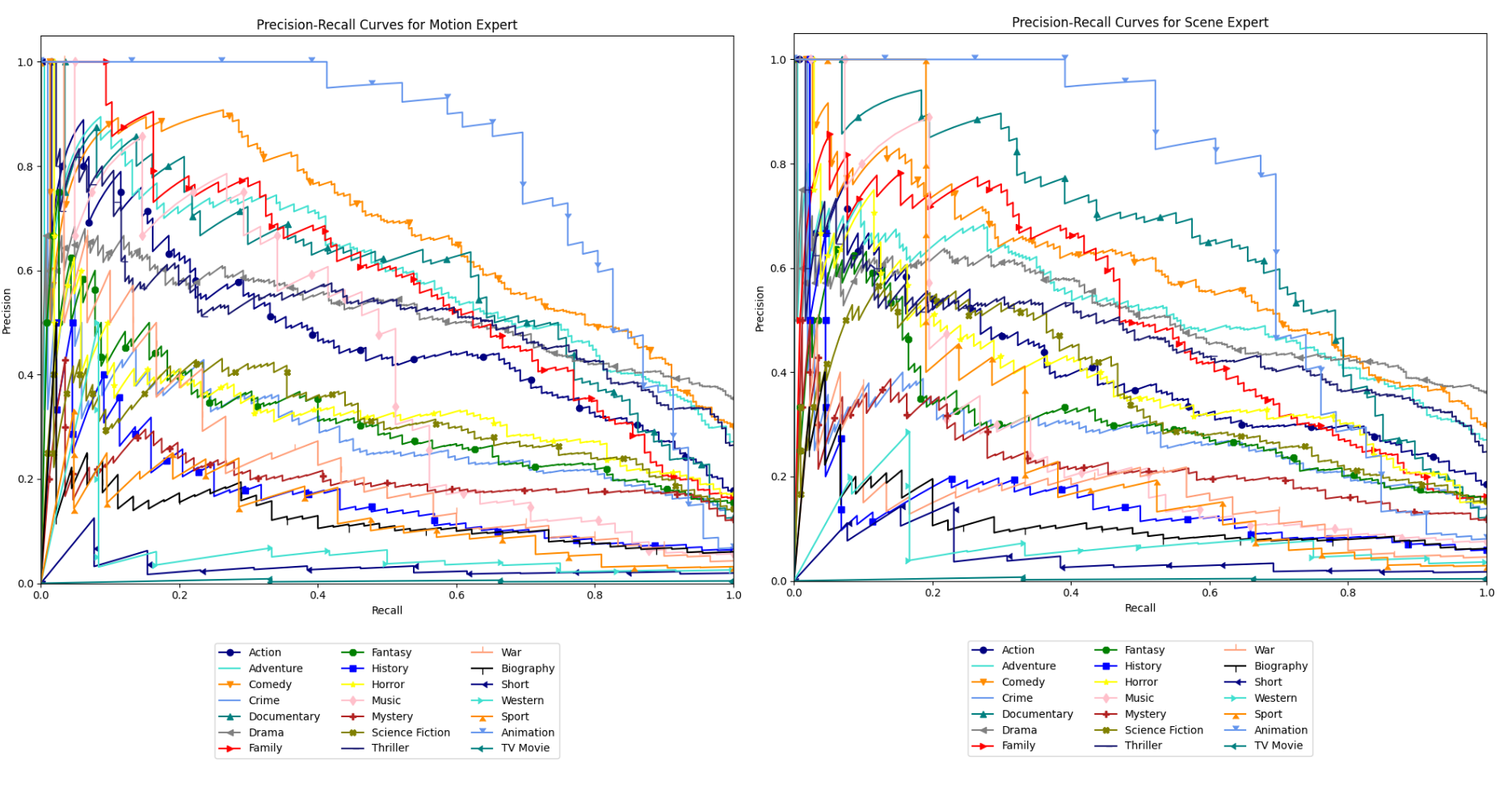}
\label{fig:recal_classes_sup  }
\includegraphics[width=1.0\textwidth]{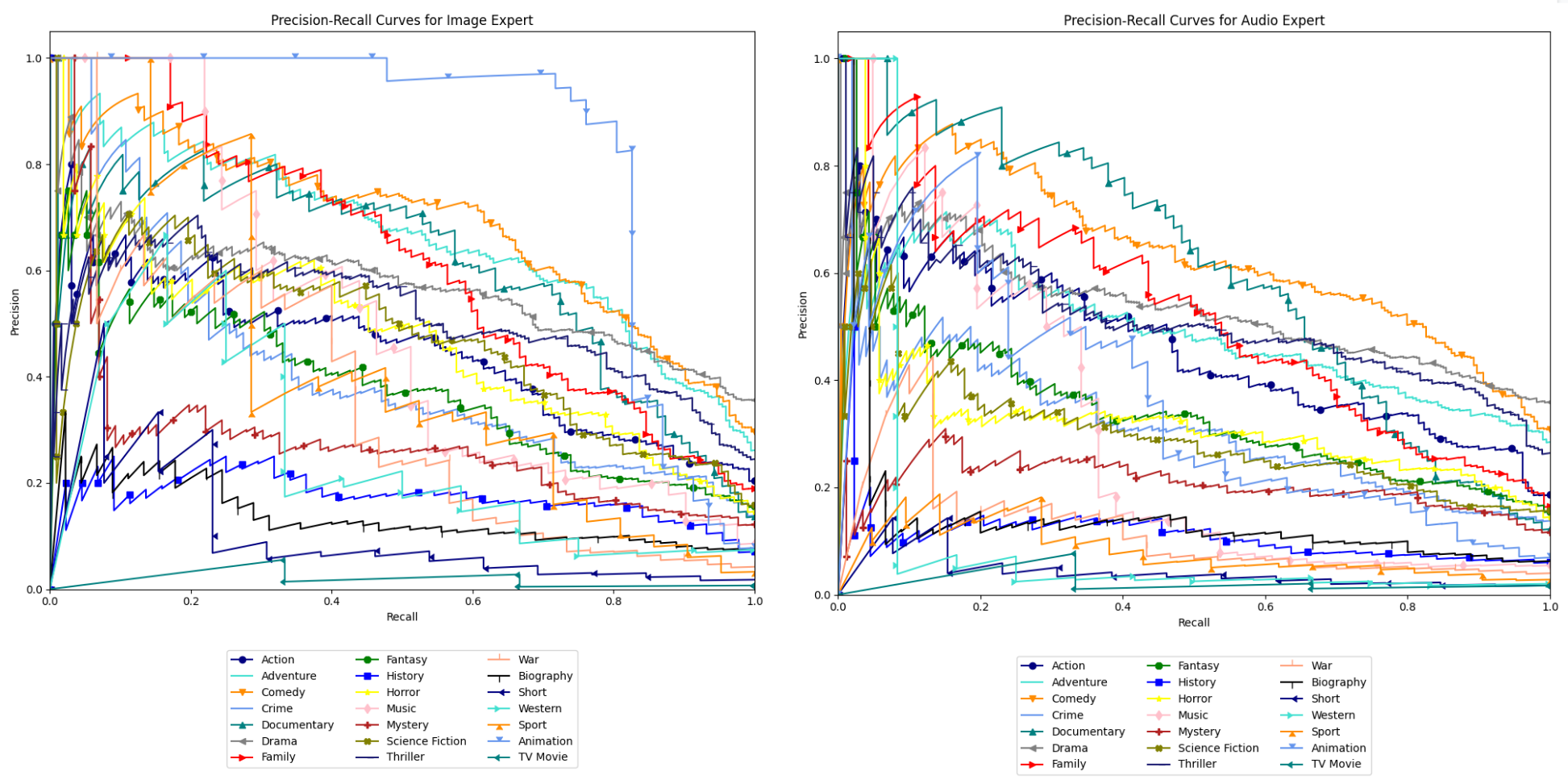}
\caption{Precision-recall curves for each expert over all labels. }
\squeezeup
\label{fig:sup_curve  }
\end{figure*}
\begin{figure*}[htp]
\centering
\includegraphics[width=1.0\textwidth]{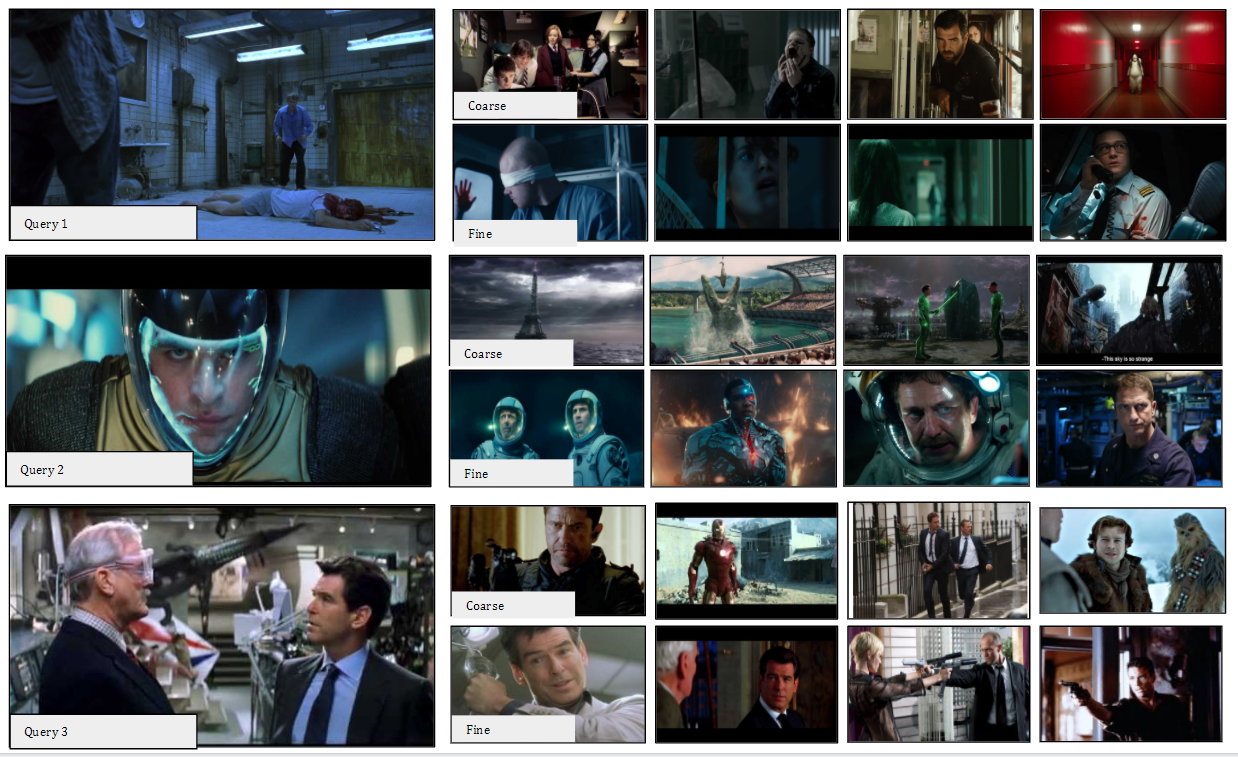}
\caption{Additional retrieval results obtained from the bottleneck embedding layer after training for coarse genre classification, and after fine-tuning with the fine-grained self-supervised network. }
\label{fig:retrieval_sup  }
\squeezeup
\end{figure*}
\begin{table}[htbp]
\centering
\caption{The effect of sequence length on classification accuracy across a number of metrics. }
\begin{tabular}{l|l|l|l|l}
Sequence Length & $F1_w$  &$(\overline{PRC})$&$P_w$     & $R_w$ \\
\midrule
1               & 0.456 & 0.451      & 0.475  &   0.484      \\
5               &   0.493    &   0.518         &   0.428     & 0.625        \\
9               &  0.564     &    0.583        &    0.554    &    0.611    \\
20             & 0.495        & 0.503       &  0.493    & 0.576 \\
\bottomrule
\end{tabular}
\label{tab:Seq_len_effect}
\end{table}

In Tab \ref{tab:Seq_len_effect} we show the effect of sequence length on classification accuracy across a number of metrics.  We find that longer sequences assist the model in making more accurate genre predictions suggesting that temporal information is captured through the concatenation of scenes. After a sequence length of 10, however, we notice that accuracy begins to decrease. We select a sequence length of 9 scenes for the model to ensure we can use as much as the dataset as possible, without compromising on performance.

In Fig \ref{fig:scene_preds} we can see that our model makes good predictions on individual scenes and offers reasonable guesses considering the content of the scene in isolation of the whole trailer. This explains why the model performs poorly with shorter sequences. We also notice how genre predictions change over the duration of a trailer on a scene by scene basis. This is even more prevalent in modern movies, where genre fluidity is more common or where other genres are referenced for narrative and stylistic effect.

\section{Effect of individual experts}

In Fig \ref{fig:sup_curve  } we show the precision-recall curves for each label and modal expert. As might be expected `Animation' is the best performing label on visual experts with `Comedy' performing well over both audio and visual experts. `Documentary' is best identified by the scene and audio experts, perhaps because of the additional dialogue and use of establishing shots in `Documentary' trailers. While we expected the audio expert to perform best on the `Music' label, we find that image and scene experts perform just as well. This may be due to the image expert identifying instruments and the scene expert associating music with auditoriums and stadiums. The influence of different experts over the labels demonstrates the advantage of using a collaboratively gated, multi-modal approach.

\section{MLP Dimensions}
To assist in implementation we provide further description of the multi-layer perceptrons used throughout the network. 
\begin{itemize}

    \item After scene detection and feature extraction, the collaborative gating module returns an L2 Normalised feature vector of dimension 1 x 128 which encompasses the scaled multi-modal information for that individual clip.
    \item As we are working with 9 clips per sequence in the implementation, we concatenate the 9 x 128 vectors to create a 1 x 1152 sequence vector. This is then passed through the $j.$ two-layer MLP with ReLU activations, and resulting in a sequence vector of size 1 x 256.
    \item Each sequence is then concatenated to produce a 1 x 10240 trailer vector. The size is a result of 40 sequences being concatenated. $k.$ represents the bottleneck layer where the 1 x 10240 embedding is processed via another MLP of two layers with ReLU activations generating a feature embedding of shape 1 x 2048. 
    \item For genre classification, the MLP $l.$ is defined by 2 layers with ReLU activation on the first layer of size 2048 x 1024. The final layer produces a logits vector of size 1 x 21. To asses the accuracy of the model during training we pass the logits to a sigmoid function and return each class activated over a threshold of 0.3.
    \item For fine-tuning the 1 x 2048 embedding vector is instead processed via two MLP's; $m.$ and $n.$ resulting in an embedding vector of shape 1 x 128. 
    \item Following training and fine-tuning, embeddings of shape 1 x 2048 are taken from the bottleneck layer and used as trailer representations to compare cosine similarity. 

\end{itemize}

\section{Conclusion}

While previous works have shown the effectiveness of convolutional neural networks and deep learning for genre classification, these methods do not address the unique inter-textual differences that exist within these discreet labels. Using a collaborative gated multi-modal network, we show that genre labels can be subdivided and extended by discovering semantic differences between the videos within these categories. By first training a genre classifier, rather than training a self-supervised model from scratch, we encourage the video embeddings to retain genre specific information. Continuing to train unsupervised introduces contextual awareness of the multi-modal content, and augments the embedding to closer align with similar movies while retaining genre specific information. We hope that the ability to cluster videos in this way will assist in the retrieval and classification of film styles, and benefit researchers working in film theory, archiving, and video recommendation.

\clearpage
{\small
\bibliographystyle{ieee_fullname}
\bibliography{AGRef, EFRef}
}

\end{document}